\documentclass[10pt,twocolumn,letterpaper]{article}

\usepackage{iccv}
\usepackage{times}
\usepackage{epsfig}
\usepackage{graphicx}
\usepackage{amsmath}
\usepackage{amssymb}
\usepackage{makecell}
\usepackage{multirow}

\usepackage[dvipsnames]{xcolor}
\usepackage[pagebackref=true,breaklinks=true,letterpaper=true,colorlinks,bookmarks=false]{hyperref}

\iccvfinalcopy 

\def\iccvPaperID{7920} 

\ificcvfinal\fi

\begin{document}

\title{Vision Transformer with Progressive Sampling}

\author{Xiaoyu Yue$^{*1}$ \quad Shuyang Sun$^{*2}$ \quad Zhanghui Kuang$^3$ \quad Meng Wei$^4$ \quad \\
Philip Torr$^2$\quad Wayne Zhang$^{3, 6}$\quad Dahua Lin$^{1, 5}$\\
$^1$Centre for Perceptual and Interactive Intelligence \quad
$^2$University of Oxford \\
$^3$SenseTime Research \quad
$^4$Tsinghua University \quad
$^5$The Chinese University of Hong Kong \\
$^6$Qing Yuan Research Institute, Shanghai Jiao Tong University}

\maketitle
\ificcvfinal\thispagestyle{empty}\fi

\let\thefootnote\relax\footnotetext{$^*$Equal Contribution.}

\begin{abstract}
Transformers with powerful global relation modeling abilities have been introduced to fundamental computer vision tasks recently.
As a typical example, the Vision Transformer (ViT) directly applies a pure transformer architecture on image classification, by simply splitting images into tokens with a fixed length, and employing transformers to learn relations between these tokens.
However, such naive tokenization could destruct object structures, assign grids to uninterested regions such as background, and introduce interference signals. 
To mitigate the above issues, in this paper, we propose an iterative and progressive sampling strategy to locate discriminative regions.
At each iteration, embeddings of the current sampling step are fed into a transformer encoder layer, and a group of sampling offsets is predicted to update the sampling locations for the next step.
The progressive sampling is differentiable. When combined with the Vision Transformer, the obtained PS-ViT network can adaptively learn where to look.
The proposed PS-ViT is both effective and efficient.
When trained from scratch on ImageNet, PS-ViT performs 3.8\% higher than the vanilla ViT in terms of top-1 accuracy with about $4\times$ fewer parameters and $10\times$ fewer FLOPs.
Code is available at \url{https://github.com/yuexy/PS-ViT}.
\end{abstract}

\section{Introduction}

Transformers~\cite{Vaswani2017,devlin2018bert} have become the de-facto standard architecture for natural language processing tasks. Thanks to their powerful global relation modeling abilities, researchers attempt to introduce them to fundamental computer vision tasks such as image classification~\cite{Chen2020a,Touvron,Dosovitskiy2021,Wu2020,Ramachandran2019}, object detection~\cite{zhu2021deformable,Carion2020,Zheng2020,Dai2020,Sun2020} and image segmentation~\cite{Wang2020} recently. However, transformers are initially tailored for processing mid-size sequences, and of quadratic computational complexity \wrt the sequence length. Thus, they cannot directly be used to process images with massive pixels.

\begin{figure}
    \begin{center}
    \includegraphics[width=\linewidth]{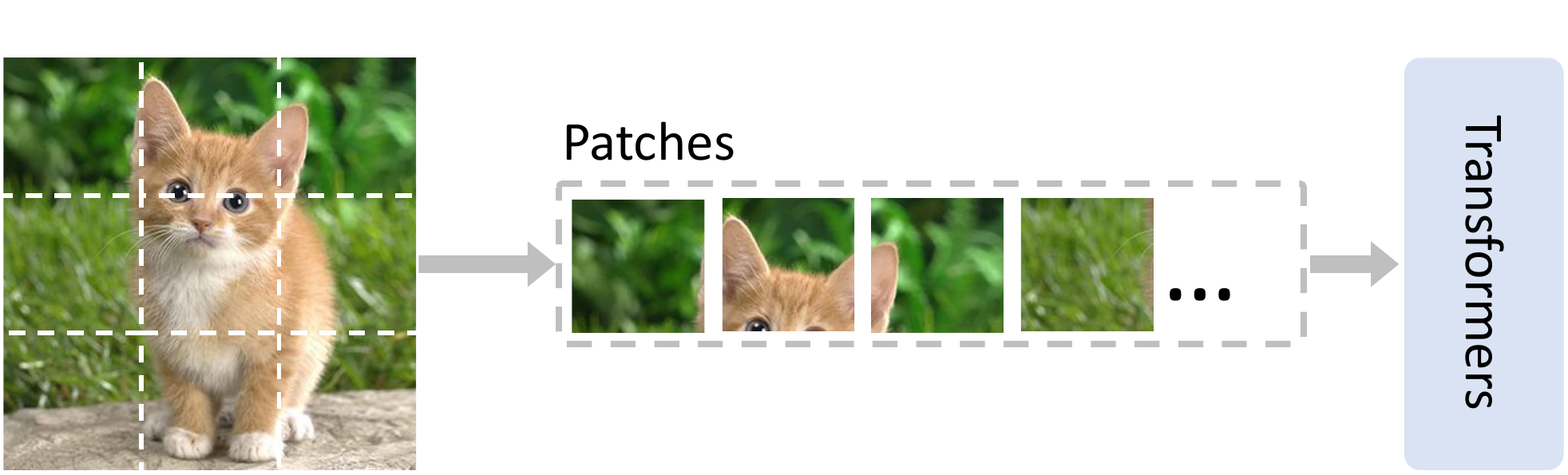}
    (a)
    \includegraphics[width=\linewidth]{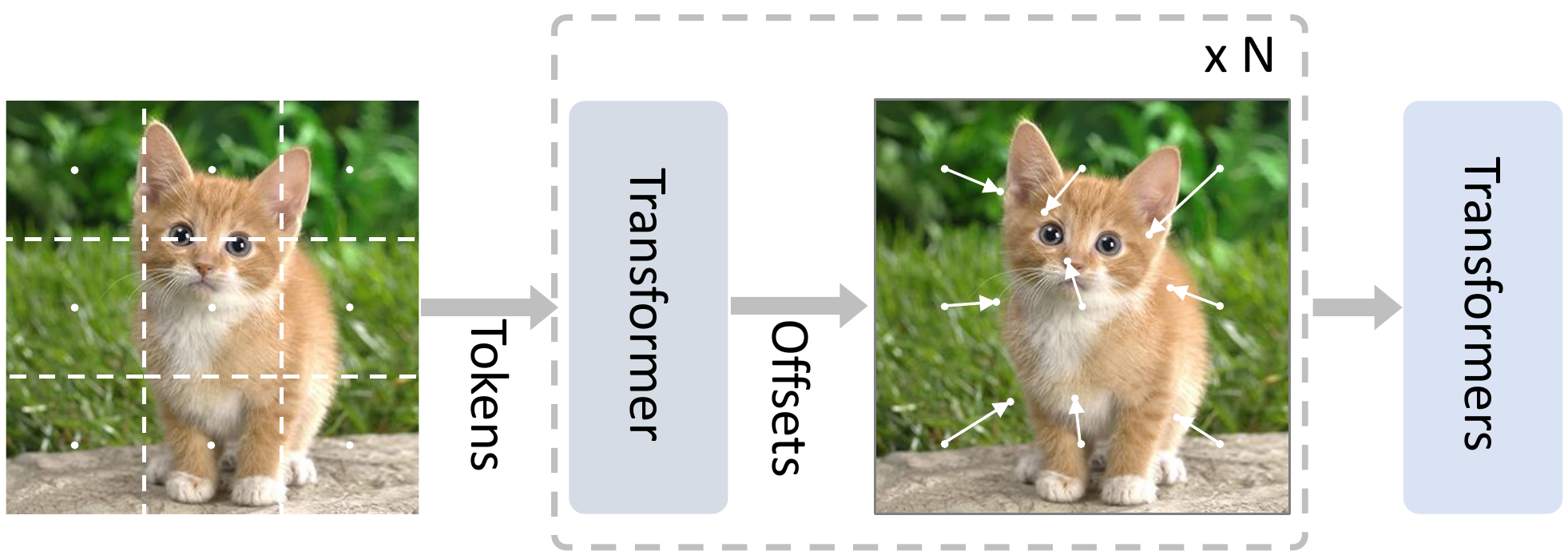}
    (b)
    \end{center}
    \caption{Comparison between the naive tokenization scheme in ViT~\cite{Dosovitskiy2021} and the progressive sampling in our proposed PS-ViT. (a) The naive tokenization scheme generates a sequence of image patches which are embedded and then fed into a stack of transformers. (b) Our PS-ViT iteratively samples discriminative locations. $\times N$ indicates $N$ sampling iterations.}
    \label{fig:fig_1}
\end{figure}

\begin{figure*}
    \centering
    \includegraphics[width=\linewidth]{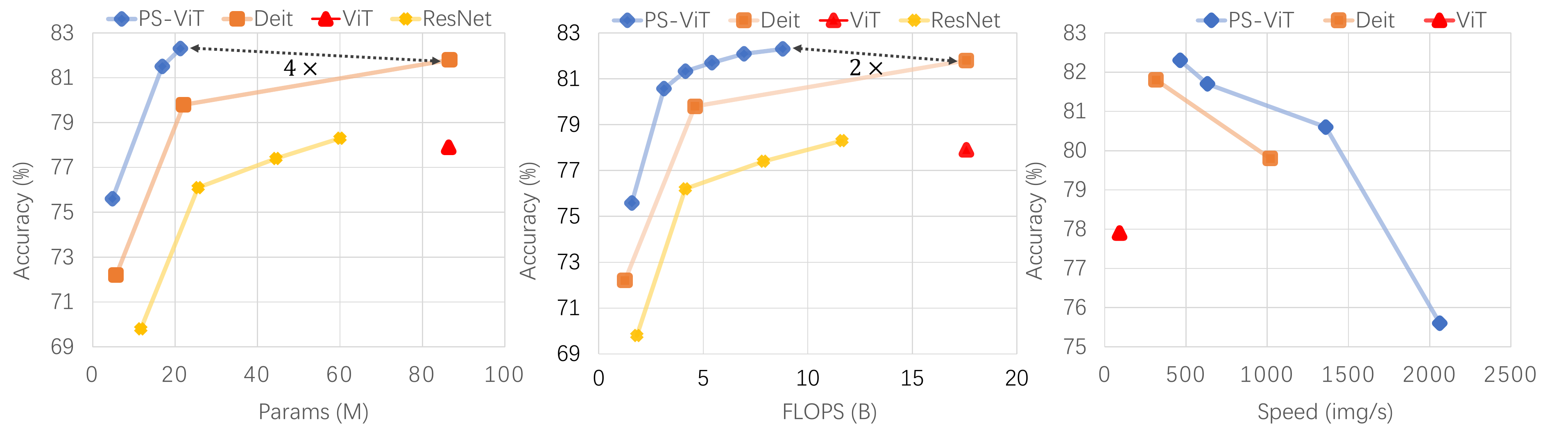}
    \caption{Comparisons between PS-ViT with  state-of-the-art networks in terms of top-1 accuracy on ImageNet, parameter number, FLOPs, and speed. The chart on the left, middle and right show top-1 accuracy \vs parameter numbers, FLOPs and speed respectively. The speed is tested on the same V100 with a batch size of 128 for fair comparison.}
    \label{fig:fig_2}
\end{figure*}

To overcome the computational complexity issue, the pioneer Vision Transformer (ViT)~\cite{Dosovitskiy2021} adopts a naive tokenization scheme that partitions one image into a sequence of regularly spaced patches, which are linearly projected into tokens.
In this way, the image is converted into hundreds of visual tokens, which are fed into a stack of transformer encoder layers for classification.
ViT attains excellent results, especially when pre-trained on large-scale datasets, which proves that full-transformer architecture is a promising alternative for vision tasks. However, the limitations of such a naive tokenization scheme are obvious.
First, the hard splitting might separate some highly semantically correlated regions that should be modeled with the same group of parameters, which destructs inherent object structures and makes the input patches to be less informative.
Figure~\ref{fig:fig_1} (a) shows that the cat head is divided into several parts, resulting in recognition challenges based on one part only.
Second, tokens are placed on regular grids irrespective of the underlying image content.
Figure~\ref{fig:fig_1} (a) shows that most grids focus on the uninterested background, which might lead to the interesting foreground object is submerged in interference signals.

The human vision system organizes visual information in a completely different way than indiscriminately processing a whole scene at once. Instead, it progressively and selectively focuses attention on interesting parts of the visual space when and where it is needed while ignoring uninterested parts, combining information from different fixations over time to understand the scene~\cite{doi:10.1080/135062800394667}.

Inspired by the procedure above, we propose a novel transformer-based progressive sampling module, which is able to learn where to look in images, to mitigate the issues caused by the naive tokenization scheme in ViT~\cite{Dosovitskiy2021}.
Instead of sampling from fixed locations, our proposed module updates the sampling locations in an iterative manner.
As shown in Figure~\ref{fig:fig_1} (b), at each iteration, tokens of the current sampling step are fed into a transformer encoder layer, and a group of sampling offsets is predicted to update the sampling locations for the next step.
This mechanism utilizes the capabilities of the transformer to capture global information to estimate offsets towards regions of interest, by combining with the local contexts and the positions of current tokens.
In this way, attention progressively converges to discriminative regions of images step by step as what human vision does.
Our proposed progressive sampling is differentiable, and readily plugged into ViT instead of the hard splitting, to construct end-to-end Vision Transforms with Progressive Sampling Networks dubbed as PS-ViT.  Thanks to task-driven training, PS-ViT tends to sample object regions correlated with semantic structures. Moreover, it pays more attention to foreground objects while less to ambiguous background compared with simple tokenization.

The proposed PS-ViT outperforms the current state-of-the-art transformer-based approaches when trained from scratch on ImageNet.
Concretely, it achieves $82.3\%$ top-1 accuracy on ImageNet which is 
higher than that of the recent ViT's variant DeiT~\cite{Touvron} with only about $4\times$ fewer parameters and $2\times$ fewer FLOPs. As shown in Figure~\ref{fig:fig_2}, we observe that PS-ViT is remarkably better, faster, and more parameter-efficient compared to state-of-the-art transformer-based networks ViT and DeiT.

\section{Related Work}

Transformers are first proposed for sequence models such as machine translation~\cite{Vaswani2017}. Benefiting from their powerful global relation modeling abilities, and highly efficient training, transformers achieve significant improvements and become the de-facto standard in many Natural Language Processing (NLP) tasks~\cite{devlin2018bert,NEURIPS2020_1457c0d6,Peters2018,peters2019knowledge,yang2019xlnet}.

\noindent \textbf{Transformers in Computer Vision.}
Inspired by the success of transformers in NLP tasks, many researchers attempt to apply transformers, or attention mechanism in computer vision tasks, such as image classification~\cite{Chen2020a,Touvron,Dosovitskiy2021,Wu2020,Ramachandran2019,Bello2019,local_relation_huhan,sun2019fishnet}, object detection~\cite{zhu2021deformable,Carion2020,Zheng2020,Dai2020,guided_cnn,Sun2020}, image segmentation~\cite{Wang2020}, low-level image processing~\cite{Chen2020,Yang2020,Parmar2018}, video understanding~\cite{Xia2020} generation~\cite{Weissenborn2020}, multi-modality understanding~\cite{Chen2020b,Sun2019,li2019visualbert}, and Optical Character Recognition (OCR) \cite{wang2019simple,yue2020robustscanner,sheng2019nrtr}. Transformers' powerful modelling capacity comes at the cost of computational complexity. Their consumed memory and computation grow quadratically \wrt the token length, which prevents them to being directly applied to images with massive pixels as tokens.
Axial attention~\cite{Ho2019} applied attention along a single axis of the tensor without flattening to reduce the computational resource requirement. iGPT~\cite{Chen2020a} simply down-sampled images to one low resolution, trained a sequence of transformers to auto-regressively predict pixels and achieved promising performance with a linear probe.
ViT~\cite{Dosovitskiy2021} regularly partitioned one high-resolution image into $16\times 16$ patches, which were fed into one pure transformer architecture for classification, and attained excellent results even compared to state-of-the-art convolutional networks for the first time.
However, ViT needs pretraining on large-scale datasets, thereby limiting their adoption.
DeiT~\cite{Touvron} proposed a data-efficient training strategy and a teacher-student distillation mechanism~\cite{hinton2015distilling}, and improved ViT's performance greatly. Moreover, it is trained on ImageNet only, and thus considerably simplifies the overall pipeline of ViT.
Our proposed PS-ViT also starts from ViT. Instead of splitting pixels into a small number of visual tokens, we propose a novel progressive sampling strategy to avoid structure destruction and focus more attention on interesting regions.

\noindent \textbf{Hard Visual Attention.} PS-ViT as a series of glimpses akin to hard visual attention~\cite{Mnih2014,Ba2015a,Xu2015,Elsayed2019}, makes decisions based on a subset of locations only in the input image. However, PS-ViT is differentiable and can be easily trained in an end-to-end fashion while previous hard visual attention approaches are non-differentiable and trained with Reinforcement Learning (RL) methods. These RL-based methods have proven to be less effective when scaled onto more complex datasets~\cite{Elsayed2019}. Moreover, our PS-ViT targets at progressively sampling discriminative tokens for Vision Transformers while previous approaches locate interested regions for convolutional neural networks~\cite{Mnih2014,Ba2015a,Elsayed2019} or sequence decoders~\cite{Xu2015}.
%
Our work is also related to the deformable convolution \cite{dai2017deformable, deformablev2} and deformable attention \cite{deformable_detr} mechanism, however, the motivation and the way of pixel sampling in this work are different from what proposed in the deformable convolution and attention mechanism.

\section{Methodology}

In this section, we first introduce our progressive sampling strategy
and then describe the overall architecture of our proposed PS-ViT network.
Finally, we will elaborate on the details of PS-ViT.
Symbols and notations of our method are presented in Table~\ref{tab:symbol_list}.

\begin{table}[]
    \begin{center}
    \begin{tabular}{l|l}
        \Xhline{3\arrayrulewidth}
        Name & Description \\
        \hline
        $\mathbf{F}\in\mathbb{R}^{C\times H\times W}$ & \makecell[l]{the feature map extracted by the \\feature extractor module}\\
        \hline
        $\mathbf{p}_t\in\mathbb{R}^{2\times(n\times n)}$ & \makecell[l]{the sampling points at the iteration $t$}\\
        \hline
        $\mathbf{P}_t\in\mathbb{R}^{C\times(n\times n)}$ & \makecell[l]{the position embeddings at the \\iteration $t$}\\
        \hline
        $\mathbf{o}_t\in\mathbb{R}^{2\times(n\times n)}$ & \makecell[l]{the sampling offsets at the \\iteration $t$}\\
        \hline
        $\mathbf{T}_t^{'}\in\mathbb{R}^{C\times(n\times n)}$ & \makecell[l]{tokens sampled from $\mathbf{F}$ at \\the iteration $t$}\\
        \hline
        $\mathbf{T}_t\in\mathbb{R}^{C\times(n\times n)}$ & \makecell[l]{tokens predicted by the progressive \\sampling module at the iteration $t$}\\
        \hline
        $C$ & \makecell[l]{the dimension of tokens}\\
        \hline
        $N$ & \makecell[l]{the iteration number in the \\progressive sampling module}\\
        \hline
        $N_v$ & \makecell[l]{the transformer layer number in the \\vision transformer module}\\
        \hline
        $M$ & \makecell[l]{the head number}\\
        \Xhline{3\arrayrulewidth}
    \end{tabular}
    \end{center}
    \caption{The list of symbols and notations used in this paper.}
    \label{tab:symbol_list}
\end{table}

\subsection{Progressive Sampling}
ViT~\cite{Dosovitskiy2021} regularly partitions one image into $16\times 16$ patches, which are linearly projected into a set of tokens, regardless of the content importance of image regions and the integral structure of objects.
To pay more attention to interesting regions of images and mitigate the problem of structure destruction,  we propose one novel progressive sampling module. As it is differentiable,  it is adaptively driven via the following vision transformer based image classification task.
\begin{figure}
    \begin{center}
    \includegraphics[width=\linewidth]{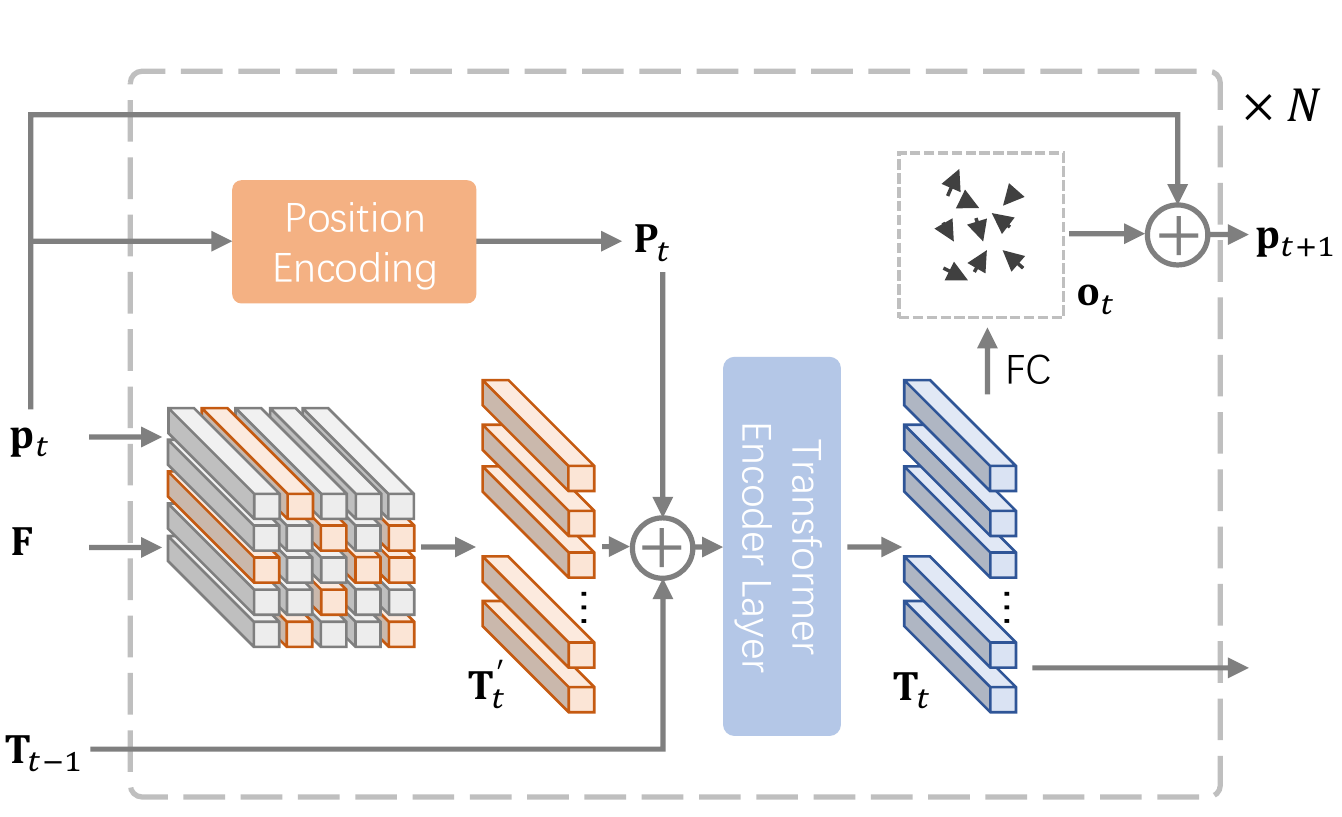}
    \end{center}
    \caption{The architecture of the progressive sampling module. At each iteration, given the sampling location $\mathbf{p}_t$ and the feature map $\mathbf{F}$, we sample the initial tokens $\mathbf{T}_t^{'}$  at $\mathbf{p}_t$ over $\mathbf{F}$, which are   element-wisely added with the positional encodings $\mathbf{P}_t$ generated based on $\mathbf{p}_t$, and the output tokens $\mathbf{T}_{t-1}$ of the last iteration, and then fed into one transformation encoder layer to predict the tokens $\mathbf{T}_t$ of the current iteration. The offset matrix $\mathbf{o}_t$ is predicted via one fully-connected layer based on $\mathbf{T}_t$, which is added with $\mathbf{p}_t$ to obtain the sampling positions $\mathbf{p}_{t+1}$ for the next iteration. The above procedure is iterated $N$ times.}
    \label{fig:ps_module}
\end{figure}

Our progressive sampling module is an iterative framework. Given the input feature map $\mathbf{F}\in \mathbb{R}^{C\times H\times W}$ with $C$, $H$, and $W$ being the feature channel dimension, height and width respectively, it finally outputs a sequence of tokens $\mathbf{T}_N\in \mathbb{R}^{C\times (n\times n)}$, where $(n\times n)$ indicates the number of samples over one image and $N$ is the total iterative number in the progressive sampling module.

As shown in Figure~\ref{fig:ps_module}, at each iteration, the sampling locations are updated via adding them with the offset vectors of the last iteration. Formally,
\begin{equation}
     \mathbf{p}_{t+1} = \mathbf{p}_{t}+\mathbf{o}_{t}, \; t\in \{1,\dots, N-1\},
     \label{eq:offset}
\end{equation}
where $\mathbf{p}_t\in \mathbb{R}^{2\times (n\times n)}$, and $\mathbf{o}_{t}\in \mathbb{R}^{2\times (n\times n)}$ indicate the sampling location matrix and the offset matrix predicted at the iteration $t$. For the first iteration, we initialize the $\mathbf{p}_1$ to be the regularly-spaced locations as done in ViT~\cite{Dosovitskiy2021}.  Concretely, the $i$-th location $\mathbf{p}_1^i$ is given by
\begin{equation}
    \begin{split}
        \mathbf{p}_1^i&= [\pi_i^ys_h+{s_h}/{2},\pi_i^xs_w+{s_w}/{2}]\\
    \pi_i^y &= \left \lfloor i/n\right \rfloor\\
    \pi_i^x &= i-\pi_i^y*n\\
    s_h &= {H}/{n} \\
    s_w &= {W}/{n},
    \end{split}
\end{equation}
where $\pi_i^y$ and $\pi_i^x$ map the location index $i$ to the row index and the column one respectively.
$\left \lfloor \cdot \right \rfloor$ indicates the floor operation. $s_h$ and $s_w$ are the step size in the $y$ and $x$ axial direction respectively. Initial tokens are then sampled over the input feature map at the sampled locations as follows:
\begin{equation}
     \mathbf{T}_t^{'} = \mathbf{F}(\mathbf{p}_t), \; t\in \{1,\dots, N\},
     \label{eq:sampling}
\end{equation}
where $\mathbf{T}_t^{'}\in \mathbb{R}^{C\times (n\times n)}$ is the initial sampled tokens at the iteration $t$.
As elements of $\mathbf{p}_t$ are fractional, the sampling is implemented via the bilinear interpolation operation, which is differentiable \wrt both the input feature map $\mathbf{F}$ and the sampling locations $\mathbf{p}_t$.  The initial sampled tokens, the output tokens of the last iteration, and the positional encodings of the current sampling locations are further element-wisely added before being fed into one transformer encoder layer to get the output tokens of the current iteration.
Formally, we have
\begin{equation}
    \begin{split}
        \mathbf{P}_{t} &= \mathbf{W}_t \mathbf{p}_t \\
        \mathbf{X}_t &= \mathbf{T}_t^{'} \oplus \mathbf{P}_{t} \oplus \mathbf{T}_{t-1} \label{eq:eq_fusion} \\
        \mathbf{T}_t &= \text{Transformer}(\mathbf{X}_t), \; t\in \{1,\dots, N\},
    \end{split}
\end{equation}
where $\mathbf{W}_t\in \mathbb{R}^{C\times 2}$ is the linear transformation that projects the sampled locations $\mathbf{p}_t$ to the positional encoding matrix $\mathbf{P}_{t}$ of size $C\times (n\times n)$, all iterations share the same $\mathbf{W}_t$.
$\oplus$ indicates the element-wise addition while $\text{Transformer}(\cdot)$ is the mulit-head self-attention based transformer encoder layer, which will be elaborated in Section~\ref{sec:sec_details}. Note that $\mathbf{T}_0$ is a zero matrix in Equation~ (\ref{eq:eq_fusion}). 
ViT~\cite{Dosovitskiy2021} uses the 2-D sinusoidal positional embeddings of patch indices. Since their patches are regularly spaced,
patch indices can exactly encode the relative coordinates of patch centers in one image. However, it does not hold true in our case as our sampled locations are non-isometric as shown in Figure~\ref{fig:fig_1}. We project the normalized absolute coordinates of sampled locations to one embedding space as the positional embeddings instead.
Finally, the sampling location offsets are predicted for the next iteration except at the last iteration as follows:
\begin{equation}
\mathbf{o}_{t}=\mathbf{M}_t\mathbf{T}_t, \; t\in \{1,\dots, N-1\},
\end{equation}
where $\mathbf{M}_t\in \mathbb{R}^{2\times C}$ is the learnable linear transformation for predicting the sampling offset matrix. 

With the progressive sampling strategy, the sampled locations progressively converge to interesting regions of images. Therefore, we name it by progressive sampling.

\subsection{Overall Architecture}

\begin{figure*}
    \begin{center}
    \includegraphics[width=\linewidth]{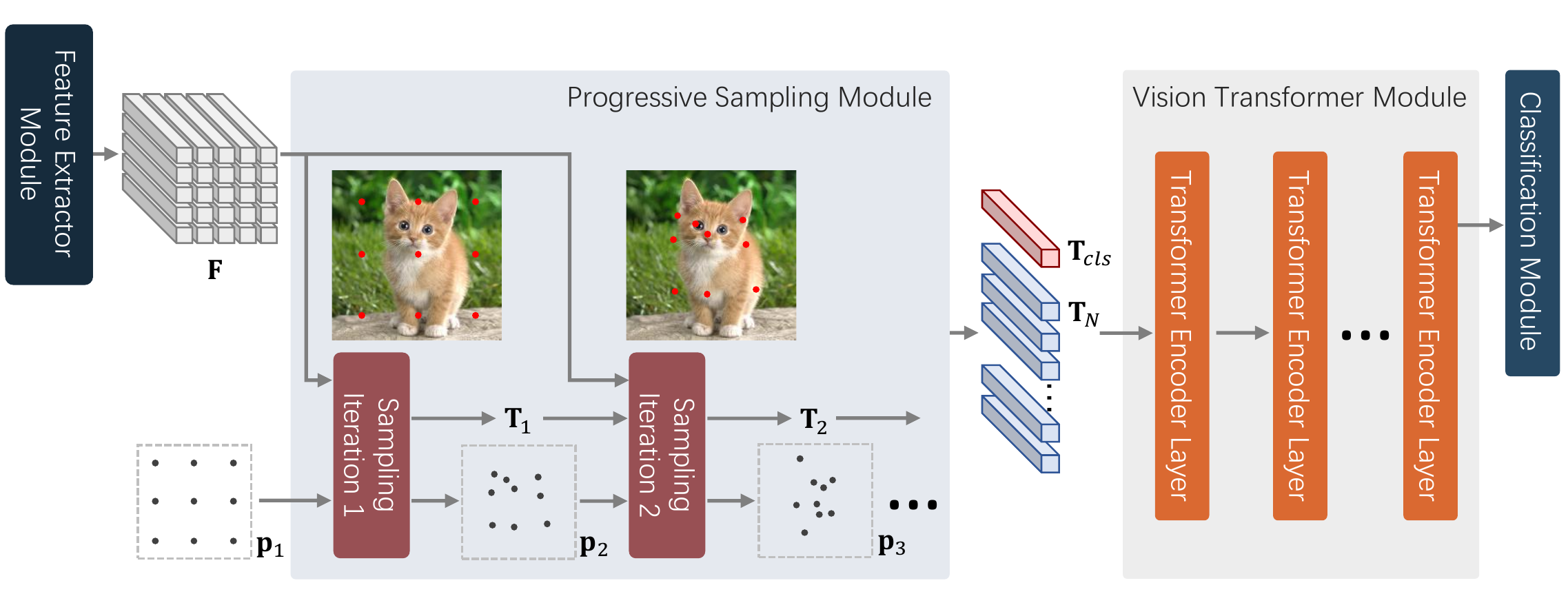}
    \end{center}
    \caption{Overall architecture of the proposed Progressive Sampling Vision Transformer (PS-ViT). Given an input image, its feature map $\mathbf{F}$ is first extracted by the feature extractor module. Tokens $\mathbf{T}_i$ are then sampled progressively and iteratively at adaptive locations $\mathbf{p}_i$ over $\mathbf{F}$ in the progressive sampling module. The final output tokens $\mathbf{T}_{N}$ of the progressive sampling module  are padded with the classification token $\mathbf{T}_{cls}$ and further fed into the vision transformer module to refine $\mathbf{T}_{cls}$, which is finally classified in the classification module.} 
    \label{fig:arch}
\end{figure*}

As shown in Figure~\ref{fig:arch}, the architecture of the PS-ViT consists of four main components: 1) a feature extractor module to predict dense tokens; 2) a progressive sampling module to sample discriminative locations; 3) a vision transformer module that follows the similar configuration of ViT~\cite{Dosovitskiy2021} and DeiT~\cite{Touvron}; 4) a classification module.
 
The feature extractor module aims at extracting the dense feature map $\mathbf{F}$, where the progressive sampling module can simple tokens $\mathbf{T}_t$.
Each pixel of the dense feature map $\mathbf{F}$ can be treated as a token associated with a patch of the image.
We employ the convolutional stem and the first two residual blocks in the first stage of the ResNet50~\cite{he2016deep} as our feature extractor module because the convolution operator is especially effective at modeling spatially local contexts.

The vision transformer module follows the architecture adopted in ViT~\cite{Dosovitskiy2021} and DeiT~\cite{Touvron}.
We pad an extra token named by the classification token $\mathbf{T}_{cls} \in \mathbb{R}^{C\times 1}$ to the output tokens $\mathbf{T}_N$ of the last iteration in the progressive sampling module, and fed them into the vision transformer module.
Formally,
\begin{equation}
    \overline{\mathbf{T}} = \text{VTM}(\left [ \mathbf{T}_{cls}, \mathbf{T}_N \right ]),
\end{equation}
where $\text{VTM}$ indicates the vision transformer module function which is a stack of transformer encoder layers, and $\overline{\mathbf{T}}\in \mathbb{R}^{C\times (n\times n+1)}$ is the output.
Note that, the positional information has been fused into $\mathbf{T}_N$ in the progressive sampling module, so we do not need to add positional embedding here.
The classification token refined through the vision transformer module is finally used to predict the image classes. We use the cross entropy loss to end-to-end train the proposed PS-ViT network.

\subsection{Implementation}
\label{sec:sec_details}

\begin{table}[]
    \begin{center}
    \begin{tabular}{c|c|c|c|c|c|c}
        \Xhline{3\arrayrulewidth}
         Networks& \makecell[c]{$N$} &  \makecell[c]{$N_v$} &$C$ &M& \#params & \#FLOPs\\
        \hline
        PS-ViT-Ti & 4 & 8 & 192 & 3 & 4.7 M & 1.6 B\\
        PS-ViT-Ti$^{\dagger}$ & 4 & 8 & 192 & 3 & 3.6 M & 1.6 B\\
        PS-ViT-B & 4 & 10 & 384 & 6 & 21.3 M & 5.4 B\\
        PS-ViT-B$^{\dagger}$ & 4 & 10 & 384 & 6 & 16.9 M & 5.4 B\\
        \Xhline{3\arrayrulewidth}
    \end{tabular}
    \end{center}
    \caption{PS-ViT configurations. ${\dagger}$ indicates weight sharing between different iterations in the Progressive Sampling Module (PSM). $N$, $N_v$, $C$ and $M$ are the iteration number in PSM, the transformer encoder layer number in the vision transformer module, the dimension of tokens, and the head number in each transformer respectively.}
    \label{tab:tab_config}
\end{table}

\noindent \textbf{Transformer Encoder Layer.}
The transformer encoder layer serves as the basic building block for the progressive sampling module and the vision transformer module.
Each transformer encoder layer has a \textit{multi-head self-attention} and a \textit{feed-forward unit}.

Given queries $\mathbf{Q} \in\mathbb{R}^{D\times L}$, keys $\mathbf{K}\in \mathbb{R}^{D\times L}$ and values $\mathbf{V} \in \mathbb{R}^{D\times L}$ with $D$ being the dimension and $L$ being the sequence length, the scaled dot-product self attention can be calculated as:
\begin{equation}
    \text{Attn}(\mathbf{Q}, \mathbf{K}, \mathbf{V}) =  \text{softmax}(\mathbf{Q}^{T}\mathbf{K} / \sqrt{D})\mathbf{V}^{T},
\end{equation}
where $\mathbf{Q}^{T}$ indicates the transpose of $\mathbf{Q}$, and $\text{softmax}(\cdot)$ is the softmax operation applied over each row of the input matrix.
For Multi-Head self-Attention (MHA), the queries, keys and values are generated via linear transformations on the inputs for $M$ times with one individual learned weight for each head.
Then attention function is applied in parallel on  queries, keys and values of each head.
Formally,
\begin{equation}
\begin{aligned}
    \text{MHA}(\mathbf{Z}) &= \mathbf{W}^o[\mathbf{H}_1,\dots, \mathbf{H}_M]^{T}, \\
    \mathbf{H}_i &= \text{Attn}(\mathbf{W}^Q_i\mathbf{Z}, \mathbf{W}^K_i\mathbf{Z}, \mathbf{W}^V_i\mathbf{Z}),
\end{aligned}
\end{equation}
where $\mathbf{W}^o\in \mathbb{R}^{\frac{C}{M}\times C}$ is a learnable linear projection. $\mathbf{W}^Q_i\in \mathbb{R}^{\frac{C}{M}\times C}$, $\mathbf{W}^K_i\in \mathbb{R}^{\frac{C}{M}\times C}$ and $\mathbf{W}^V_i\in \mathbb{R}^{\frac{C}{M}\times C}$ are the linear projections for the queries, keys and values of the $i$-th head respectively. 

The feed-forward unit of the transformer encoder layer consists of two fully connected layers with one GELU non-linear activation~\cite{hendrycks2016gaussian} between them and the latent variable dimension being $3C$. For simplicity, the transformer encoder layers in both the progressive sampling module and the vision transformer module keep the same settings.

\noindent \textbf{Progressive Sampling Back-propagation.}
The back-propagation of the progressive sampling is straightforward.
According to Equation (\ref{eq:offset}) and Equation (\ref{eq:sampling}), for each sampling location $i$, the gradient \wrt the sampling offsets $\mathbf{o}_t^i$ at the iteration $t$ is computed as:

\begin{equation}
\begin{aligned}
    \frac{\partial \mathbf{T}_t^{' i}}{\partial \mathbf{o}_{t-1}^{i}} &= \frac{\partial \mathbf{F}(\mathbf{p}_{t-1}^{i} + \mathbf{o}_{t-1}^{i})}{\partial \mathbf{o}_{t-1}^{i}} \\
    & = \sum_{\mathbf{q}} \frac{\partial K(\mathbf{q}, \mathbf{p}_{t-1}^{i} + \mathbf{o}_{t-1}^{i})}{\partial \mathbf{o}_{t-1}^{i}} \mathbf{F}(\mathbf{q}),
\end{aligned}
\end{equation}
where $K(\cdot, \cdot)$ is the kernel for bilinear interpolation to calculate weights for each integral spatial location $\mathbf{q}$.

\begin{table}[htbp]
    \begin{center}
    \setlength{\tabcolsep}{1.6pt}
    \scalebox{1.0}{
    \begin{tabular}{l|c|c|c|c|c}
        \Xhline{3\arrayrulewidth}
        Model & \makecell[c]{Image\\size} & \makecell[c]{Params\\ (M)} & \makecell[c]{FLOPs \\(B)}  & \makecell[c]{Top-1\\ (\%)} & \makecell[c]{Top-5\\ (\%)} \\
        \Xhline{3\arrayrulewidth}
        \multicolumn{6}{c}{\textbf{CNN-based}} \\
        \hline
        R-18~\cite{he2016deep} & 224$^2$ & 11.7 & 1.8 & 69.8 & 89.1 \\
        R-50~\cite{he2016deep} & 224$^2$ & 25.6 & 4.1 & 76.1 & 92.9 \\
        R-101~\cite{he2016deep} & 224$^2$ & 44.5 & 7.9 & 77.4 & 93.5 \\
        \hline
        X-50-32$\times$4d~\cite{xie2017aggregated} & 224$^2$ & 25.0 & 4.3 & 79.3 & 94.5 \\
        X-101-32$\times$4d~\cite{xie2017aggregated} & 224$^2$ & 44.2 & 8.0 & 80.3 & 95.1 \\
        \hline
        RegNetY-4GF~\cite{Radosavovic2020} & 224$^2$ & 20.6 & 4.0 & 79.4 & - \\
        RegNetY-6.4GF~\cite{Radosavovic2020} & 224$^2$ & 30.6 & 6.4 & 79.9 & - \\
        RegNetY-16GF~\cite{Radosavovic2020} & 224$^2$ & 83.6 & 15.9 & 80.4 & - \\
        \Xhline{3\arrayrulewidth}
        \multicolumn{6}{c}{\textbf{Transformer-based}} \\
        \hline
        ViT-B/16~\cite{Dosovitskiy2021} & 384$^2$ & 86.4 & 55.5 & 77.9 & - \\
        \hline
        DeiT-Ti~\cite{Touvron} & 224$^2$ & 5.7 & 1.3 & 72.2 & - \\
        DeiT-S~\cite{Touvron} & 224$^2$ & 22.1 & 4.6 & 79.8 & - \\
        DeiT-B~\cite{Touvron} & 224$^2$ & 86.4 & 17.6 & 81.8 & - \\
        \hline
        PS-ViT-Ti/14 & 224$^2$ & 4.8 & 1.6 & 75.6 & 92.9 \\
        PS-ViT-B/10 & 224$^2$ & 21.3 & 3.1 & 80.6 & 95.2 \\
        PS-ViT-B/14 & 224$^2$ & 21.3 & 5.4 & 81.7 & 95.8 \\
        PS-ViT-B/18 & 224$^2$ & 21.3 & 8.8 & 82.3 & 96.1 \\
        \Xhline{3\arrayrulewidth}
    \end{tabular}
    }
    \end{center}
    \caption{Comparison with state-of-the-art networks on ImageNet with single center crop testing. The number after ``/'' is the sampling number in each axial direction. \eg, PS-ViT-Ti/14 indicates PS-ViT-Ti with $14\times 14$ sampling locations.}
    \label{tab:tab_comp_sota}
\end{table}

\noindent \textbf{Network Configuration.}
The feature dimension $C$, the iteration number $N$ in the progressive sampling module, the vision transformer layer number $N_v$ in the vision transformer module, and the head number $M$ in each transformer layer affect the model size, FLOPs, and performances. In this paper, we configure them with different speed-performance tradeoffs in Table~\ref{tab:tab_config} so that the proposed PS-ViT can be used in different application scenarios.
The number of sampling points along each spatial dimension $n$ is set as $14$ by default.

Considering the sampling  in each iteration  is  conducted over the same feature map $\mathbf{F}$ in the progressive sampling module, 
we try to share weights between those iterations to further reduce the number of trainable parameters.
As shown in Table~\ref{tab:tab_config}, about 25$\%$ parameters can be saved in this setting.

\section{Experiments}

\begin{table}[htbp]
    \begin{center}
    \setlength{\tabcolsep}{6mm}
    \begin{tabular}{c|c}
        \Xhline{3\arrayrulewidth}
        Epochs & 300 \\
        Optimizer & AdamW \\
        Batch size & 512\\
        Learning rate & 0.0005\\
        Learning rate decay & cosine\\
        Weight decay & 0.05\\
        Warmup epochs & 5\\
        \hline
        Label smooth & 0.1 \\
        Dropout & 0.1 \\
        Rand Augment & (9, 0.5) \\
        Mixup probability & 0.8 \\
        CutMix probability & 1.0 \\
        \Xhline{3\arrayrulewidth}
    \end{tabular}
    \end{center}
    \caption{Training strategy and hyper-parameter settings.}
    \label{tab:strategy}
\end{table}

\subsection{Experimental Details on ImageNet}

All the experiments for image classification are conducted on the ImageNet 2012 dataset~\cite{krizhevsky2012imagenet} that includes 1k classes, 1.2 million images for training, and 50 thousand images for validation.
We train our proposed PS-ViT on ImageNet without pretraining on large-scale datasets.
We train all the models of PS-ViT using PyTorch~\cite{pytorch} with 8 GPUs.
Inspired by the data-efficient training as done in~\cite{Touvron}, we use the AdamW~\cite{Loshchilov2019} as the optimizer. The total training epoch number and the batch size are set to $300$ and $512$ respectively. The learning rate is initialized with $0.0005$, and decays with the cosine annealing schedule~\cite{sgdr_cosine}. We regularize the loss via the smoothing label with $\epsilon=0.1$. We use random crop, Rand-Augment~\cite{Cubuk2020}, Mixup\cite{Zhang2018a}, and CutMix~\cite{Yun2019} to augment images during training. Images are resized to $256\times 256$, and cropped at the center with $224\times 224$ size when testing.
Training strategy and its hyper-parameter settings are summarized in Table~\ref{tab:strategy}.

\subsection{Results on ImageNet}
We compare our proposed PS-ViT with state-of-the-art networks on the standard image classification benchmark ImageNet in terms of parameter numbers, FLOPS, and top-1 and top-5 accuracies in Table~\ref{tab:tab_comp_sota}.

\noindent \textbf{Comparison with CNN based networks.} 
Our PS-ViTs considerably outperform ResNets~\cite{he2016deep} while with much fewer parameters and FLOPs.
Specifically, Compared with ResNet-18, PS-ViT-Ti/14 absolutely improves the top-1 accuracy by 5.8\% while reducing 6.9 M parameters and 0.2 B FLOPs. We can observe a similar trend when comparing PS-ViT-B/10 (PS-ViT-B/14) and ResNet-50 (ResNet-101).
Our proposed PS-ViT achieves superior performance and computational efficiency when compared with the state-of-the-art CNN based network RegNet~\cite{Radosavovic2020}. Particularly, when compared with RegNetY-16GF, PS-ViT-B/18 improves the top-1 accuracy by 1.8\% with about a quarter of parameters and a half of FLOPS.

\noindent \textbf{Comparison with transformer based networks.}
Table~\ref{tab:tab_comp_sota} shows that our proposed PS-ViT outperforms ViT~\cite{Dosovitskiy2021} and its recent variant DeiT~\cite{Touvron}.
In particular, PS-ViT-B/18 achieves 82.3\% top-1 accuracy which is 0.5\% higher than the baseline model DeiT-B while with 21 M parameters and 8.8 B FLOPs only.
Our performance gain attributes to two parts. First, PS-ViT samples CNN-based tokens which is more efficient than raw image patches used in ViT~\cite{Dosovitskiy2021} and DeiT~\cite{Touvron}. Second, our progressive sampling module can adaptively focus on regions of interest and produce more semantically correlated tokens than the naive tokenization used in~\cite{Dosovitskiy2021,Touvron}.

\subsection{Ablation Studies}
The PS-ViT models predict on the class token in all the ablation studies.

\noindent \textbf{A larger sampling number $n$ leads to better performance.}
We first evaluate how the sampling number parameter $n$ affects the PS-ViT performance. The sequence length of sampled tokens which is fed into the vision transformer module is $n^2$. The more the sampled tokens, the more information PS-ViT can extract. However, sampling more tokens would increase the computation and memory usage. Table~\ref{tab:tab_sampling_num} reports the FLOPs, and top-1 and top-5 accuracies with different $n$. It has been shown that the FlOPs increases as $n$ becomes larger, and the accuracy increases when $n\le16$ and 
plateaus when $n>16$. Considering the speed-accuracy trade-off, we set $n=14$ by default except as otherwise noted.
\begin{table}[t]
    \begin{center}
    \scalebox{0.96}{
    \begin{tabular}{c|c|c|c|c}
        \Xhline{3\arrayrulewidth}
        $n$ & Params (M) & FLOPs (B) & Top-1 (\%) & Top-5 (\%) \\
        \hline
        \hline
        10 & 21.3 & 3.1 & 80.6 & 95.2 \\
        12 &  & 4.2 & 81.3 & 95.5 \\
        14 &  & 5.4 & 81.7 & 95.8 \\
        16 &  & 7.0 & 82.1 & 95.8 \\
        18 &  & 8.8 & 82.3 & 96.1 \\
        \Xhline{3\arrayrulewidth}
    \end{tabular}}
    \end{center}
    \caption{Effect of the sampling number $n$ in each axial direction.}
    \label{tab:tab_sampling_num}
\end{table}

\begin{table}[t]
    \begin{center}
    \begin{tabular}{c|c|c|c}
        \Xhline{3\arrayrulewidth}
        Model & $N$ & Top-1 (\%) & Top-5 (\%) \\
        \hline
        \hline
        PS-ViT-B/14 & 1 & 80.6 & 95.3 \\
         & 2 & 81.5 & 95.6 \\
         & 4 & 81.7 & 95.8 \\
         & 6 & 81.8 & 95.7 \\
         & 8 & 81.9 & 95.7 \\
         & 9 & 81.7 & 95.8 \\
         & 10 & 81.6 & 95.8 \\
        \Xhline{3\arrayrulewidth}
    \end{tabular}
    \end{center}
    \caption{Effect of the iteration number $N$ in the progressive sampling module.}
    \label{tab:tab_iteration_num}
\end{table}

\noindent \textbf{The performance can be further improved with more iterations of progressive sampling.}
We then evaluate the effect of the iteration number $N$ of the progressive sampling module in Table~\ref{tab:tab_iteration_num}. To keep the computational complexity unchanged, all models in Table~\ref{tab:tab_iteration_num} have $14-N$ transformer layers in the vision transformer module, and totally $14$ transformer layers in the entire network.
$N=1$ indicates the sampling points will not be updated.
It has been shown that PS-ViT performs the best when $N=8$ and the accuracy begins to decline when $N>8$.
As we keep the total number of transformer layers unchanged, increasing $N$ will result in the decrease of transformer layers in lateral modeling, which might damage the performance.
Considering the accuracy improvement is negligible from $N=4$ to $N=8$, we set $N=4$ by default except as otherwise noted.

\noindent \textbf{Fair comparison with ViT.}
The network hyper-parameters in the transformer encoder of PS-ViT are different from the original setting of ViT. For a fair comparison, we further study how ViT performs when the network hyper-parameters are set to be the same as ours.
We set the number of layers, channels, heads, and the number of tokens to be the same as what was proposed in PS-ViT-B/14, and train the network under the same training regime.
As shown in Table~\ref{tab:comp_fair}, ViT achieves 78.4\% top-1 accuracy, which is greatly inferior to its PS-ViT counterpart.
We thereby conclude that the progressive sampling module can fairly boost the performance of ViT.

\begin{table}[t]
    \begin{center}
    \setlength{\tabcolsep}{3mm}
    \begin{tabular}{c|c|c}
        \Xhline{3\arrayrulewidth}
         & Top-1 (\%) & Top-5 (\%) \\
        \hline
        \hline
        ViT$*$ & 78.4 & 94.1 \\
        PS-ViT-B/14& 81.7 & 95.8 \\
        \Xhline{3\arrayrulewidth}
    \end{tabular}
    \end{center}
    \caption{Comparison between our PS-ViT with ViT. $*$ means the model with the same model configuration and training strategy.}
    \label{tab:comp_fair}
\end{table}

\begin{figure*}
    \begin{center}
    \begin{minipage}{0.19\linewidth}
        \centerline{ \includegraphics[height=\linewidth, width=\linewidth]{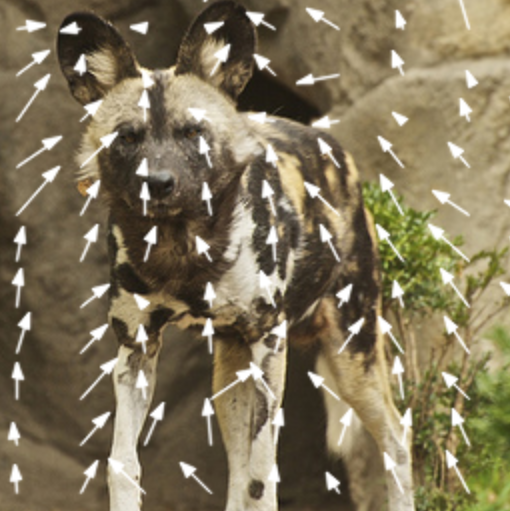}}
    \end{minipage}
    \begin{minipage}{0.19\linewidth}
        \centerline{ \includegraphics[height=\linewidth, width=\linewidth]{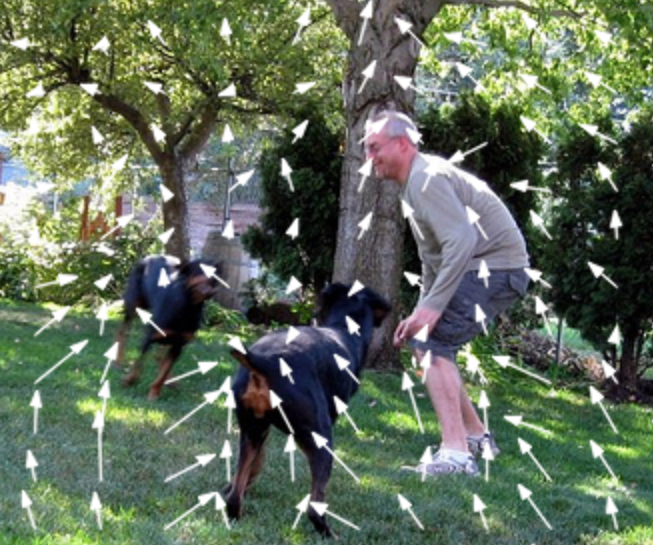}}
    \end{minipage}
    \begin{minipage}{0.19\linewidth}
        \centerline{ \includegraphics[height=\linewidth, width=\linewidth]{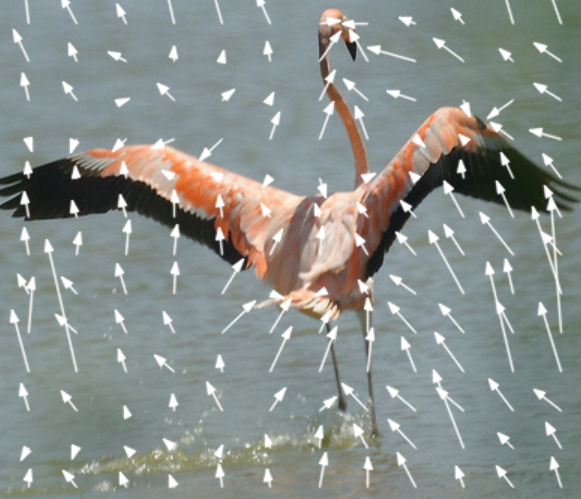}}
    \end{minipage}
    \begin{minipage}{0.19\linewidth}
        \centerline{ \includegraphics[height=\linewidth, width=\linewidth]{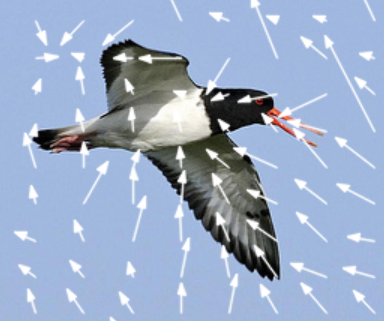}}
    \end{minipage}
    \begin{minipage}{0.19\linewidth}
        \centerline{ \includegraphics[height=\linewidth, width=\linewidth]{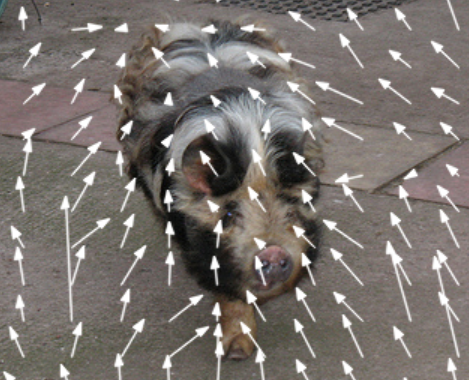}}
    \end{minipage}
    \begin{minipage}{0.19\linewidth}
        \centerline{ \includegraphics[height=\linewidth, width=\linewidth]{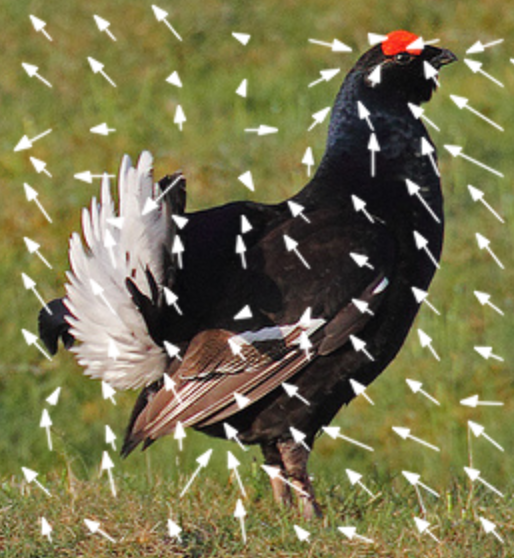}}
    \end{minipage}
    \begin{minipage}{0.19\linewidth}
        \centerline{ \includegraphics[height=\linewidth, width=\linewidth]{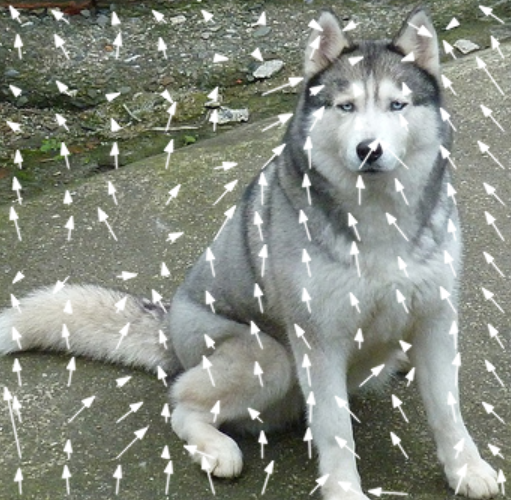}}
    \end{minipage}
    \begin{minipage}{0.19\linewidth}
        \centerline{ \includegraphics[height=\linewidth, width=\linewidth]{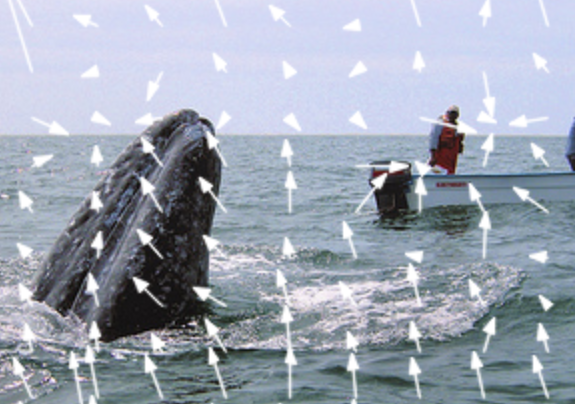}}
    \end{minipage}
     \begin{minipage}{0.19\linewidth}
        \centerline{ \includegraphics[height=\linewidth, width=\linewidth]{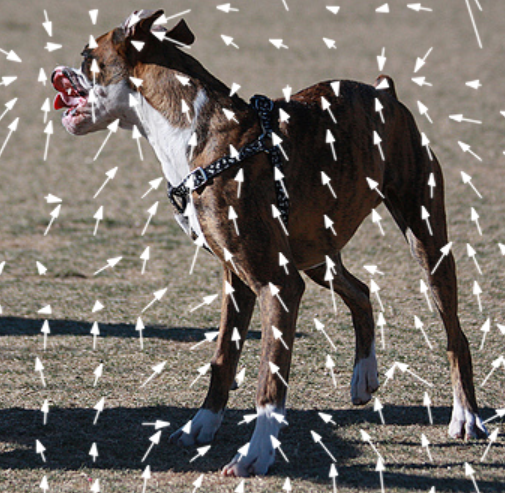}}
    \end{minipage}
     \begin{minipage}{0.19\linewidth}
        \centerline{ \includegraphics[height=\linewidth, width=\linewidth]{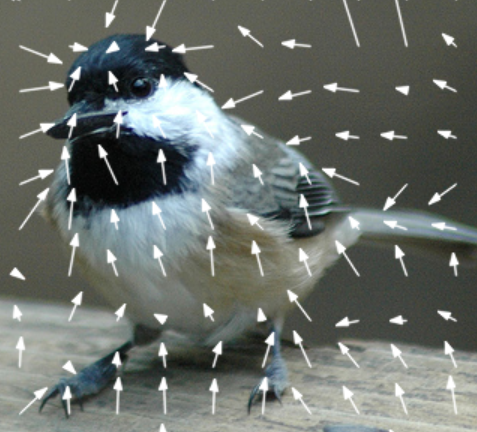}}
    \end{minipage}
    \end{center}
    \caption{Visualization of sampled locations in the proposed progressive sampling module. The start points of arrows are initial sampled locations ($\mathbf{p}_1$) while the end points of arrows are the final sampled locations ($\mathbf{p}_4$).  }
    \label{fig:visualization}
\end{figure*}

\noindent \textbf{Sharing weights between sampling iterations.}
Model size (parameter number) is one of the key factors when deploying deep models on terminal devices.
Our proposed PS-ViT is very terminal device friendly as it can share weights in the progressive sampling module with a negligible performance drop. Table~\ref{tab:tab_share} compares PS-ViT with and without weight sharing in the progressive sampling module. It has been shown that weight sharing can reduce the parameter number by about 21\%$\sim$23\% while with a slight performance drop, especially for PS-ViT-B/12 and PS-ViT-B/14.

\begin{table}[t]
    \begin{center}
    \begin{tabular}{l|c|c|c}
        \Xhline{3\arrayrulewidth}
        Model & Params (M) & Top-1 (\%) & Top-5 (\%) \\
        \hline
        \hline
        PS-ViT-Ti/14 & 4.8 & 75.6 & 92.9 \\
        PS-ViT-Ti$^{\dagger}$/14 & 3.7 & 74.1 & 92.3 \\
        \hline
        PS-ViT-B/10 & 21.3 & 80.6 & 95.2 \\
        PS-ViT-B$^{\dagger}$/10 & 16.9 & 80.0 & 94.8 \\
        \hline
        PS-ViT-B/12 & 21.3 & 81.3 & 95.5 \\
        PS-ViT-B$^{\dagger}$/12 & 16.9 & 80.9 & 95.3 \\
        \hline
        PS-ViT-B/14 & 21.3 & 81.7 & 95.8 \\
        PS-ViT-B$^{\dagger}$/14 & 16.9 & 81.5 & 95.6\\
        \Xhline{3\arrayrulewidth}
    \end{tabular}
    \end{center}
    \caption{Comparison PS-ViT with and without weight sharing in the progressive sampling module. ${\dagger}$ indicates weight sharing.}
    \label{tab:tab_share}
\end{table}

\subsection{Speed Comparison}

Our proposed PS-ViT is efficient not only in theory but also in practice. 
Table~\ref{tab:tab_speed} compare the efficiency of state-of-the-arts
networks in terms of FLOPs and speed (images per second).
For fair comparison, we measure the speed of all of the models on a server with one 32GB V100 GPU.
The batch size is fixed to 128 and the number of images that can be inferred per second is reported averaged over 50 runs.
It has been shown that PS-ViT is much more efficient than ViT and DeiT when their top-1 accuracies are comparable.
Specifically, PS-ViT-B/14 and DeiT-B have similar accuracy around 81.7\%. However, PS-ViT-B/14 achieves about 2.0 times and 3.3 times as fast as DeiT-B in terms of  speed and FLOPs respectively. PS-ViT-B/10 speeds up ViT-B/16 by about 13.7 times and 16.9 times in terms of speed and FLOPs while improving 2.7\% top-1 accuracy.

\begin{table}[]
    \begin{center}
    \begin{tabular}{l|c|c|c}
    \Xhline{3\arrayrulewidth}
        Model & FLOPs (B) & Speed (img/s) & Top-1 \\
        \hline
        \hline
        RegNetY-4.0GF & 4.0 & 1097.6 & 79.4 \\
        RegNetY-6.4GF & 6.4 & 487.0 & 79.9 \\
        RegNetY-16GF & 15.9 & 351.0 & 80.4 \\
        \hline
        ViT-B/16 & 55.5 & 92.4 & 77.9 \\
        \hline
        DeiT-S & 4.6 & 1018.2 & 79.8 \\
        DeiT-B & 17.6 & 316.1 & 81.8 \\
        \hline
        PS-ViT-Ti/14 & 1.6 & 1955.3 & 75.6 \\
        PS-ViT-B/10 & 3.1 & 1348.0 & 80.6 \\
        PS-ViT-B/14 & 5.4 & 765.6 & 81.7 \\
        PS-ViT-B/18 & 8.8 & 463.8 & 82.3 \\
        \Xhline{3\arrayrulewidth}
    \end{tabular}
    \end{center}
    \caption{Comparison the efficiency
    of PS-ViT, and that of state-of-the-art networks in terms of FLOPs and speed.}
    \label{tab:tab_speed}
\end{table}

\subsection{Visualization}
In order to explore the mechanism of the learnable sampling locations in our method, we visualize the predicted offsets of our proposed progressive sampling module in Figure~\ref{fig:visualization}.
We can observe that the sampling locations are adaptively adjusted according to the content of the images.
Sampling points around objects tend to move to the foreground area and converge to the key parts of objects.
With this mechanism, discriminative regions such as the chicken head are sampled densely, retaining the intrinsic structure information of highly semantically correlated regions.

\subsection{Transfer Learning}
\begin{table}[t]
    \centering
    \begin{tabular}{c|c|c|c|c|c}
        \Xhline{3\arrayrulewidth}
        Model & IM & C10 & C100 & Flowers & Cars \\
        \hline
        ViT-B/16 & 77.9 & 98.1 & 87.1 & 89.5 & - \\
        ViT-L/16 & 76.5 & 97.9 & 86.4 & 89.7 & - \\
        DeiT-B & 81.8 & 99.1 & 90.8 & 98.4 & 92.1 \\
        PS-ViT-B/14 & 81.7 & 99.0 & 90.8 & 98.8 & 92.9 \\
        \Xhline{3\arrayrulewidth}
    \end{tabular}
    \caption{Top-1 accuracy on other datasets. ImageNet and CIFAR are abbreviated to ``IM" and ``C".}
    \label{tab:transfer}
\end{table}

In addition to ImageNet, we also transfer PS-ViT to downstream tasks to demonstrate its generalization ability. We follow the practice done in DeiT~\cite{Touvron} for fair comparison.
Table~\ref{tab:transfer} shows results for models that have been pre-trained on ImageNet and finetuned for other datasets including CIFAR-10~\cite{krizhevsky2009learning}, CIFAR-100~\cite{krizhevsky2009learning}, Flowers-102~\cite{nilsback2008automated} and Stanford Cars~\cite{krause20133d}.
PS-ViT-B/14 can perform on-par-with or even better than DeiT-B with about $4\times$ fewer FLOPS and parameters on all these datasets, which demonstrates the superiority of our PS-ViT.

\section{Conclusions}
In this paper, we propose an efficient Vision Transformers with Progressive Sampling (PS-ViT). PS-ViT first extracts feature maps via a feature extractor, and then progressively selects discriminative tokens with one progressive sampling module. The sampled tokens are fed into a vision transformer module and the classification module for image classification.  PS-ViT mitigates the structure destruction issue in the ViT and adaptively focuses on interesting regions of objects. It achieves considerable improvement on ImageNet compared with ViT and its recent variant DeiT. 
We also provide a deeper analysis of the experimental results to investigate the effectiveness of each component. Moreover, PS-ViT is more efficient than its transformer based competitors both in theory and in practice.

\noindent \textbf{Acknowledgement.} 
This work was partially supported by Innovation and Technology Commission of the Hong Kong Special Administrative Region, China (Enterprise Support Scheme under the Innovation and Technology Fund B/E030/18).

{\small
\bibliographystyle{ieee_fullname}
\bibliography{egbib}
}

\end{document}